\documentclass[letterpaper]{article} 
\usepackage[preprint]{aaai2027}  
\usepackage[hyphens]{url}  
\usepackage{graphicx} 
\usepackage{natbib}  
\usepackage{caption} 
\usepackage{algorithm}
\usepackage{algorithmic}

\usepackage{newfloat}
\usepackage{listings}
\DeclareCaptionStyle{ruled}{labelfont=normalfont,labelsep=colon,strut=off} 
\floatstyle{ruled}
\newfloat{listing}{tb}{lst}{}
\floatname{listing}{Listing}

\usepackage{booktabs}

\usepackage{amsmath}
\usepackage{subcaption}
\usepackage[table]{xcolor}
\definecolor{winrow}{RGB}{221,243,222}        
\definecolor{baselinerow}{RGB}{246,207,203}   
\definecolor{preRowColor}{RGB}{200,72,72}     
\definecolor{postRowColor}{RGB}{44,95,138}    
\newcommand{\best}[1]{\cellcolor{winrow}\textbf{#1}}
\newcommand{\base}[1]{\cellcolor{baselinerow}#1}
\newcommand{\preRow}[1]{\textcolor{preRowColor}{\textbf{#1}}}
\newcommand{\postRow}[1]{\textcolor{postRowColor}{\textbf{#1}}}

\title{Rethinking Normalization Placement for LLMs: \\ Post-Norm under Curriculum Depth Growing}
\author{
  Sheng Ren$^{1}$,
  Yadong Wang$^{1}$,
  Naiqiang Tan$^{2}$,
  Jiangang Kong$^{2}$,
  Jun Fang$^{2}$,\\
  Rui Liu$^{2}$,
  Jun Wang$^{2}$,
  Kai Chen$^{2}$,
  Lipeng Liang$^{2}$,
  Xiang Chen$^{1}$\corresponding 
}
\affiliations{
  $^{1}$Nanjing University of Aeronautics and Astronautics, Nanjing, Jiangsu, China\\
  $^{2}$Didichuxing Co. Ltd, Beijing, China\\
  \texttt{\{rensheng,xiang\_chen\}@nuaa.edu.cn}
}

\begin{document}

\maketitle

\begin{abstract}
Pre-norm is the standard normalization placement in modern Transformers because it facilitates joint optimization of full-depth models.
We ask whether this preference persists when depth is introduced through a curriculum.
In curriculum depth growth, each appended block receives the boundary representation produced by a trained prefix, making normalization placement relevant to forward conditioning.
We therefore test whether placement and training curriculum interact.
In a controlled distillation study with a Qwen3-8B teacher and a nine-layer student, pre-norm and post-norm are indistinguishable under joint training, differing by $0.0004$ validation CE, while post-norm improves over pre-norm by $0.0328$ under curriculum growth, an order of magnitude larger.
A post-joint control matched by student active-layer tokens remains worse than post-grow, which rules out compute as the sole explanation.
The ranking crosses over during the curriculum: post-norm takes the lead once blocks are appended.
Single-block and freeze controls localize the ranking change to block appending rather than shallow-block quality or retraining.
Boundary diagnostics associate post-norm with stable residual scales and pre-norm with structural-token scale drift; on a fixed batch, the final pre-grow block is also nearly identity-mapped.
Together with the phase-wise crossover, these observations are consistent with boundary-scale conditioning after new blocks are appended.
The results motivate treating normalization placement and training curriculum as coupled design choices in this distillation setting.
\end{abstract}

\section{Introduction}

Pre-norm has become the standard choice in modern Transformer language models~\cite{DBLP:conf/icml/XiongYHZZXZLWL20,DBLP:journals/corr/abs-2302-13971}. This preference is well motivated in conventional end-to-end training. When a full-depth Transformer is trained from random initialization, normalization before each sub-layer improves gradient behavior at initialization and reduces the need for carefully tuned learning-rate warm-up in practice~\cite{DBLP:conf/icml/XiongYHZZXZLWL20,DBLP:conf/emnlp/LiuLGCH20}.

This finding has influenced decoder-only LLMs and distilled Transformer students~\cite{DBLP:journals/corr/abs-2302-13971,DBLP:conf/acl/ElhoushiSLHWL0A24}. However, post-norm remains competitive when its optimization difficulties are addressed~\cite{DBLP:journals/pami/WangMDHZW24,DBLP:conf/nips/DingYHZZYLZSYT21}. Normalization placement may therefore depend on the training regime rather than on the layer formulation alone, an issue we take up here.

\begin{figure}[t]
\centering
\includegraphics[width=0.78\columnwidth]{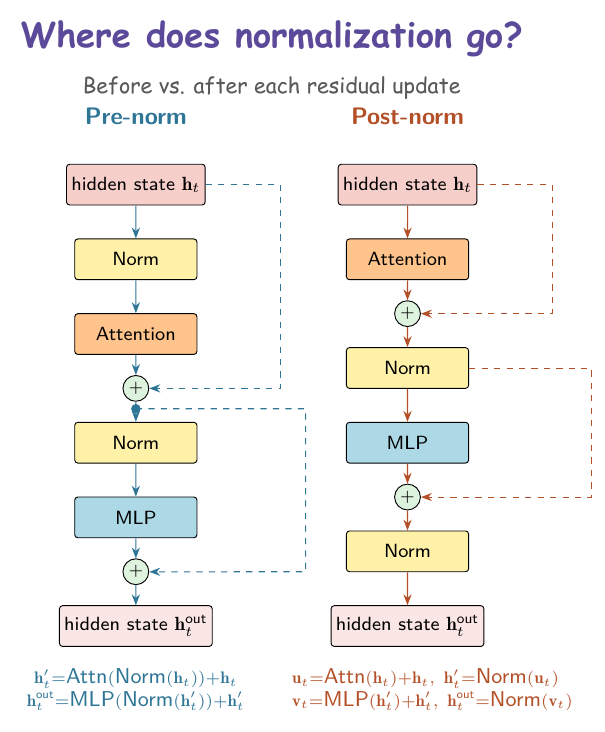}
\caption{Pre-norm normalizes sub-layer inputs, whereas post-norm normalizes the residual stream after each update.}
\label{fig:norm-compare}
\end{figure}
We study this question in curriculum depth growth. Unlike joint training, in which all layers are active from the first update, curriculum depth growth begins with a shallow model and appends new blocks in later phases~\cite{DBLP:conf/naacl/GuLYLCH21,DBLP:conf/nips/DuLQHSCG024}. The final model is therefore constructed through a sequence of inherited initializations. At each phase, the blocks trained in the previous phase determine the input distribution received by the newly appended block. This perspective is related to Net2Net-style network growth~\cite{DBLP:journals/corr/ChenGS15,DBLP:conf/icml/WeiWRC16}, but it raises a question that existing growth methods generally leave unanswered: which normalization placement provides a better-conditioned boundary input for a newly added block?

We hypothesize that normalization placement interacts with the way depth is introduced. Under joint training, normalization primarily supports optimization through a complete stack. Under curriculum depth growth, it also shapes the forward distribution passed from the trained stack to each newly appended block. Pre-norm normalizes sub-layer inputs but does not explicitly normalize the residual stream emitted across block boundaries, whereas post-norm normalizes after each residual update. We therefore expect the placements to behave similarly under joint training but to separate after the curriculum begins appending blocks.

We test this hypothesis in a controlled block-stack distillation setting. The student uses a frozen teacher embedding and a teacher-initialized language-modeling head that is trained jointly with its appendable decoder blocks. Under joint training, the two placements differ by only $0.0004$ validation CE; under curriculum growth, post-norm improves over pre-norm by $0.0328$ after additional blocks are appended, yielding an interaction of $-0.0332$. A post-joint control matched by student active-layer tokens remains worse than post-grow, which rules out additional student-decoder compute as the sole explanation. Block-boundary diagnostics show stable residual scales for post-norm and increasing scale drift for pre-norm, providing correlational evidence for the proposed mechanism.
In summary, our contributions are as follows:
\begin{itemize}
    \item We recast curriculum depth growth as block-wise initialization, shifting the normalization question from full-depth optimization to inherited boundary conditioning.

    \item We show that normalization placement interacts with the training curriculum: pre-norm and post-norm are nearly tied under joint training, whereas post-norm performs better under sequential depth growth in the evaluated block-stack setting.

    \item We analyze the interaction through single-block controls, block-removal analysis, and block-boundary activation diagnostics, obtaining converging evidence for forward conditioning.
\end{itemize}

\section{Related Work}

\paragraph{Normalization Placement in Transformers.}
The original Transformer adopted post-norm~\cite{DBLP:conf/nips/VaswaniSPUJGKP17}, while later analyses showed that pre-norm improves initialization-time gradients and reduces warm-up sensitivity~\cite{DBLP:conf/icml/XiongYHZZXZLWL20,DBLP:conf/emnlp/LiuLGCH20,DBLP:journals/corr/BaKH16}, motivating its use in modern decoder-only LLMs~\cite{DBLP:journals/corr/abs-2302-13971,DBLP:journals/corr/abs-2505-09388}.
Sandwich norm, DeepNorm, and ReZero stabilize post-norm or related residual designs~\cite{DBLP:conf/nips/DingYHZZYLZSYT21,DBLP:journals/pami/WangMDHZW24,DBLP:conf/uai/BachlechnerMMCM21}, while recent work explores placements beyond the standard dichotomy~\cite{DBLP:conf/icml/Kim0POKYSHSY25,DBLP:journals/corr/abs-2601-22095,DBLP:conf/iclr/LoshchilovHSG25}.
HybridNorm combines QKV normalization with post-norm FFNs within each block to capture the strengths of both placements~\cite{DBLP:journals/corr/abs-2503-04598}, and SiameseNorm couples pre-norm-like and post-norm-like streams through shared residual blocks to reconcile stability and representational capacity~\cite{DBLP:journals/corr/abs-2602-08064}.
A principled forward--backward stability analysis of normalization placements further clarifies why different placements induce distinct training dynamics at fixed depth~\cite{DBLP:journals/corr/abs-2510-09904}.
Initialization and residual scaling provide another stability lever~\cite{DBLP:conf/iclr/ZhangDM19,DBLP:conf/icml/HuangPBV20}.
These studies largely assume fixed-depth end-to-end training; we instead test whether the preferred placement changes when depth is introduced sequentially.
Our objective is therefore different from proposing another fixed-depth stabilization rule: we ask whether the same placement can change rank when the training path changes.

\paragraph{Network Growth and Function Preservation.}
Progressive growth expands a trained smaller network, often through function-preserving transformations~\cite{DBLP:journals/corr/ChenGS15,DBLP:conf/icml/WeiWRC16}.
Transformer variants include StackBERT, stacking operators for LLM pretraining, and block insertion into pretrained models~\cite{DBLP:conf/naacl/GuLYLCH21,DBLP:conf/nips/DuLQHSCG024,DBLP:conf/acl/WuGGLWFSL24}.
These methods primarily optimize training efficiency or preserve the function of the smaller network, while normalization placement is usually inherited and held fixed. They do not isolate whether placement itself interacts with the joint-versus-grow protocol. We instead view curriculum growth as inherited initialization: an independently parameterized block is trained on the boundary distribution produced by the existing stack, making normalization placement part of the growth condition rather than a fixed inherited choice.

\paragraph{Block-Stack Knowledge Distillation.}
Transformer distillation typically studies which logits, hidden states, or attention signals a fixed-depth student should match~\cite{DBLP:journals/corr/HintonVD15,DBLP:journals/ijcv/GouYMT21,DBLP:conf/emnlp/JiaoYSJCL0L20,DBLP:journals/corr/abs-1910-01108,DBLP:conf/acl/SunYSLYZ20,DBLP:conf/nips/WangW0B0020}.
LayerSkip shares the language-modeling head for early exits, while stochastic depth varies active layers without imposing a depth-introduction order~\cite{DBLP:conf/acl/ElhoushiSLHWL0A24,DBLP:conf/eccv/HuangSLSW16}.
Pruning methods also produce compact stacks from the top down~\cite{DBLP:conf/acl/MenXZYWL0HC25,DBLP:conf/iclr/XiaGZ024}.
These approaches determine what a fixed-depth student should retain or which existing layers should be removed, but they do not study how a trained prefix initializes the input distribution of a newly appended block. Our bottom-up block-stack setting adds independently parameterized blocks sequentially and compares normalization placements at the resulting phase boundaries, separating this question from fixed-depth distillation and top-down compression.

\begin{figure*}[t]
\centering
\includegraphics[width=\textwidth]{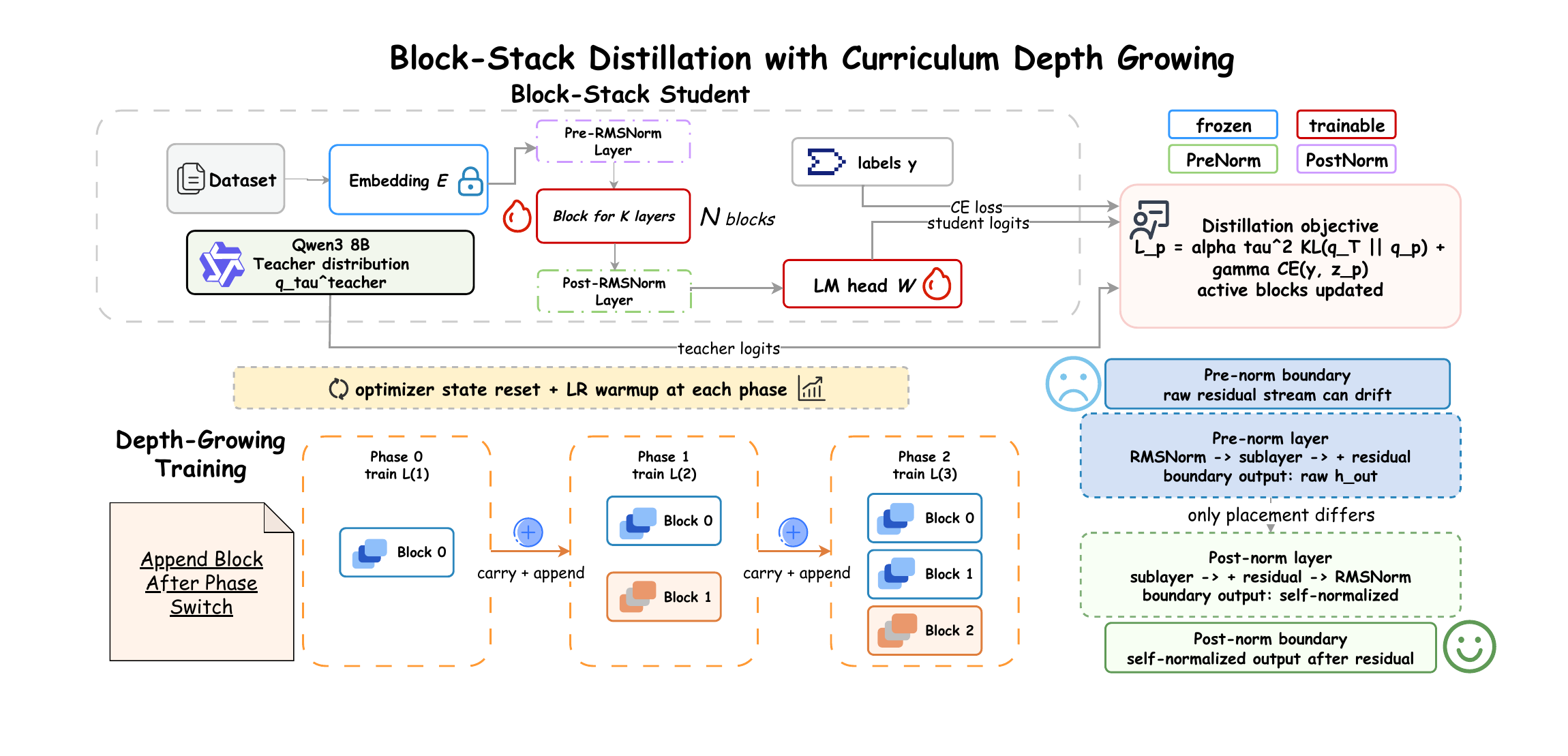}
\caption{Overview of the block-stack distillation framework.
The student uses a frozen teacher embedding and a teacher-initialized, trainable language-modeling head, while its decoder is divided into appendable blocks.
Joint training activates all blocks from the beginning, whereas curriculum growth adds one block in each phase.
Pre-norm and post-norm differ in whether normalization is applied before or after each residual update.}
\label{fig:architecture}
\end{figure*}

\section{Methodology}
\label{sec:method}

This section introduces the block-stack student, the curriculum depth-growth protocol, and the forward-conditioning hypothesis underlying our comparison.
The Experiments section describes the implementation and evaluates the interaction between normalization placement and the training curriculum. The formulation below makes the training path and block-boundary condition explicit.

\subsection{Depth-Growable Block-Stack Student}
\label{sec:student}

\paragraph{Block-Stack Student.}
We use a block-stack student whose depth can be expanded during training.
Given a teacher Transformer with $T$ decoder layers, the student is divided into $K$ blocks with $L$ decoder layers per block, resulting in a total depth of $S=K\cdot L\ll T$.
The vocabulary embedding $\mathbf{E}$ and language-modeling head $\mathbf{W}_{\mathrm{lm}}$ are copied from the teacher. The embedding remains frozen, whereas the language-modeling head and the independently parameterized student decoder layers are optimized.
This design fixes the input interface and initializes the output interface identically across variants while allowing the decoder stack and language-modeling head to adapt during distillation, consistent with common Transformer distillation settings~\cite{DBLP:journals/corr/HintonVD15,DBLP:conf/emnlp/JiaoYSJCL0L20}.
Each block contains a contiguous group of decoder layers and, once introduced, remains active in all subsequent phases.

All variants share the same attention and MLP sub-layers---rotary position embeddings~\cite{DBLP:journals/ijon/SuALPBL24}, SwiGLU activations~\cite{DBLP:journals/corr/abs-2002-05202}, grouped-query attention~\cite{DBLP:conf/emnlp/AinslieLJZLS23}, and scaled dot-product attention---so the comparison isolates the placement of RMSNorm~\cite{DBLP:conf/nips/ZhangS19a} relative to each residual update.

\paragraph{Pre-Norm and Post-Norm Layers.}
We compare two placements of RMSNorm.
Pre-norm applies RMSNorm immediately before each sub-layer but does not explicitly normalize the residual stream after each residual update:
\begin{equation}
\begin{aligned}
\mathbf{h}' &= \mathrm{Attention}(\mathrm{RMSNorm}(\mathbf{h})) + \mathbf{h}, \\
\mathbf{h}_{\mathrm{out}} &= \mathrm{MLP}(\mathrm{RMSNorm}(\mathbf{h}')) + \mathbf{h}' .
\end{aligned}
\label{eq:prenorm}
\end{equation}
Following the standard pre-norm decoder interface, we apply a final RMSNorm before the language-modeling head:
\begin{equation}
\mathbf{h}_{\mathrm{final}} = \mathbf{N}_{\mathrm{final}}(\mathbf{h}_K).
\label{eq:final_norm}
\end{equation}

Post-norm applies RMSNorm after each residual addition:
\begin{equation}
\begin{aligned}
\mathbf{u} &= \mathrm{Attention}(\mathbf{h}) + \mathbf{h},
& \mathbf{h}' &= \mathrm{RMSNorm}(\mathbf{u}), \\
\mathbf{v} &= \mathrm{MLP}(\mathbf{h}') + \mathbf{h}',
& \mathbf{h}_{\mathrm{out}} &= \mathrm{RMSNorm}(\mathbf{v}).
\end{aligned}
\label{eq:postnorm}
\end{equation}
Because the final post-norm layer already produces a normalized representation, no additional RMSNorm is applied after the final block.
The first block, however, receives the raw embedding $\mathbf{E}(\mathbf{x})$, so we apply a single entry RMSNorm $\mathbf{N}_{\mathrm{entry}}$ before it.
Both $\mathbf{N}_{\mathrm{final}}$ (pre-norm) and $\mathbf{N}_{\mathrm{entry}}$ (post-norm) are initialized from the teacher's final RMSNorm and remain trainable.
Both variants therefore contain two RMSNorm modules per decoder layer and exactly one model-boundary RMSNorm, giving them the same number of trainable normalization parameters.
Thus, internal block boundaries differ through normalization placement rather than parameter count.

\begin{algorithm}[t]
\caption{Curriculum Depth-Growth Distillation}
\label{alg:grow}
\begin{algorithmic}[1]
\REQUIRE Teacher components $(\mathbf{E}, \mathbf{N}, \mathbf{W}_{\mathrm{lm}})$, blocks $K$, layers per block $L$, loss weights $\alpha,\gamma$, and phase budgets $B_0,\dots,B_{K-1}$.
\ENSURE A trained block-stack student.
\STATE Randomly initialize all student decoder blocks.
\STATE Copy $\mathbf{E}$ and $\mathbf{W}_{\mathrm{lm}}$; freeze $\mathbf{E}$ and initialize the placement-specific boundary norm from $\mathbf{N}$.
\FOR{phase $p=0$ \TO $K-1$}
    \STATE Activate the first $p+1$ blocks.
    \STATE Reset optimizer state and restart the phase LR schedule.
    \FOR{step $t=1$ \TO $B_p$}
        \STATE Obtain teacher and active-student logits.
        \STATE Compute the KD+CE objective $\mathcal{L}_p$.
        \STATE Update all active decoder blocks, the boundary norm, and $\mathbf{W}_{\mathrm{lm}}$.
    \ENDFOR
\ENDFOR
\end{algorithmic}
\end{algorithm}

\subsection{Curriculum Depth Growth}
\label{sec:grow}

\paragraph{Growth as Inherited Initialization.}
Curriculum depth growth introduces decoder blocks in phases rather than activating the full stack from the beginning.
At phase $p$, the active student contains the first $p+1$ blocks, corresponding to $L(p+1)$ decoder layers.
The parameters learned in earlier phases are retained, and one additional block is activated when the model depth increases.
Each phase therefore initializes a deeper model from a trained prefix and a newly activated, randomly initialized block.

This protocol is related to progressive network growth and Net2Net-style expansion~\cite{DBLP:journals/corr/ChenGS15,DBLP:conf/icml/WeiWRC16,DBLP:conf/naacl/GuLYLCH21,DBLP:conf/nips/DuLQHSCG024,DBLP:conf/acl/WuGGLWFSL24}, but it does not require the expanded model to preserve the exact function of the preceding phase.
Instead, we study how the trained prefix determines the input distribution received by the newly introduced block during subsequent distillation. This boundary distribution is part of the initialization condition faced by the appended block.

\paragraph{Per-Phase Distillation Objective.}
At each phase, the active student produces
\begin{equation}
\mathbf{h}^{(p)} = \mathrm{Student}(\mathbf{x};\ \mathrm{active\_blocks}=p+1),
\label{eq:active_student}
\end{equation}
and is trained with the same distillation objective:
\begin{equation}
\mathcal{L}_p =
\alpha \tau^2
\mathrm{KL}\!\left(q_{\tau}^{\mathrm{teacher}} \,\middle\|\, q_{\tau}^{(p)}\right)
+
\gamma\,\mathrm{CE}\!\left(\mathbf{y},\mathbf{z}^{(p)}\right),
\label{eq:loss}
\end{equation}
where $q_{\tau}=\mathrm{softmax}(\mathbf{z}/\tau)$, $\tau$ is the distillation temperature, $\mathbf{z}^{(p)}$ denotes the logits of the active student, and $\mathbf{y}$ denotes the next-token labels.
All active decoder blocks, the boundary norm, and the language-modeling head are optimized jointly within each phase.

\paragraph{Joint Training as the Comparison Protocol.}
Under joint training, all $K$ blocks are active from the first optimization step and are trained end to end with the objective in Eq.~\ref{eq:loss}.
The two protocols use the same final architecture and objective but differ in staged block activation and, under the prescribed grow protocol, optimizer and learning-rate resets at phase boundaries.
The experimental setup details these protocol differences and the token- and compute-matching comparisons.

\paragraph{Placement--Curriculum Interaction.}
Let $\ell_{m,c}$ denote final held-out CE for placement $m\in\{\mathrm{pre},\mathrm{post}\}$ and curriculum $c\in\{\mathrm{joint},\mathrm{grow}\}$.
We measure the within-curriculum gap and factorial interaction as
\begin{equation}
\Delta_c=\ell_{\mathrm{post},c}-\ell_{\mathrm{pre},c},
\qquad
\mathcal{I}=\Delta_{\mathrm{grow}}-\Delta_{\mathrm{joint}}.
\label{eq:interaction}
\end{equation}
A negative $\mathcal{I}$ means that post-norm becomes relatively more favorable under curriculum growth.

\subsection{Forward-Conditioning Hypothesis}
\label{sec:why}

Curriculum depth growth creates a boundary condition specific to staged training.
When a new block is introduced, it receives the residual stream produced by the prefix trained in the preceding phase.
Normalization placement therefore affects not only computation within each layer but also the forward distribution received by each newly introduced block.

Pre-norm and post-norm handle this boundary distribution differently.
Pre-norm applies RMSNorm immediately before the attention and MLP sub-layers but does not explicitly normalize the representation after each residual update.
Post-norm applies RMSNorm after every residual update, so the representation passed to the next layer or across a block boundary is explicitly normalized.

We hypothesize that this distinction becomes important during curriculum depth growth.
Each newly introduced block is optimized on top of the boundary distribution produced by the existing stack.
Explicit normalization of block outputs may therefore provide a more stable input distribution for subsequent phases.
We test this hypothesis by comparing pre-norm and post-norm under joint training and curriculum depth growth, controlling for single-block performance and active compute, and examining activation scales and block contributions at block boundaries.

\paragraph{Testable Predictions.}
The hypothesis yields three observable predictions. First, post-norm need not outperform pre-norm before any block is appended, because the initial shallow model does not yet inherit a trained block boundary. Second, any placement gap should emerge after a depth transition rather than uniformly throughout training. Third, the placement favored by growth should exhibit better-controlled boundary representations and make nontrivial use of the appended blocks. These predictions separate forward conditioning from an unconditional post-norm advantage; the single-block, compute-matched, freeze, activation-scale, and block-removal analyses test the corresponding alternatives and observable consequences.

\section{Experiments}
\label{sec:experiments}

We conduct a series of controlled experiments to investigate whether normalization placement remains a fixed architectural choice when Transformer depth is introduced through a curriculum.
Using a block-stack distillation setting, we compare pre-norm and post-norm students under joint training and curriculum depth growth. The primary comparison matches the teacher, architecture, objective, data shard, and unique-token exposure; the curriculum protocol additionally changes active depth, optimization steps, data replay, and phase resets, with compute examined separately.
We ask whether placement changes the joint--grow ranking (\textbf{RQ1}), whether single-block quality or active compute explains the grow advantage (\textbf{RQ2}), and whether boundary dynamics support forward conditioning as a contributing mechanism (\textbf{RQ3}).

\begin{figure*}[t]
\centering
\begin{subfigure}[t]{0.48\textwidth}
\centering
\includegraphics[width=\linewidth]{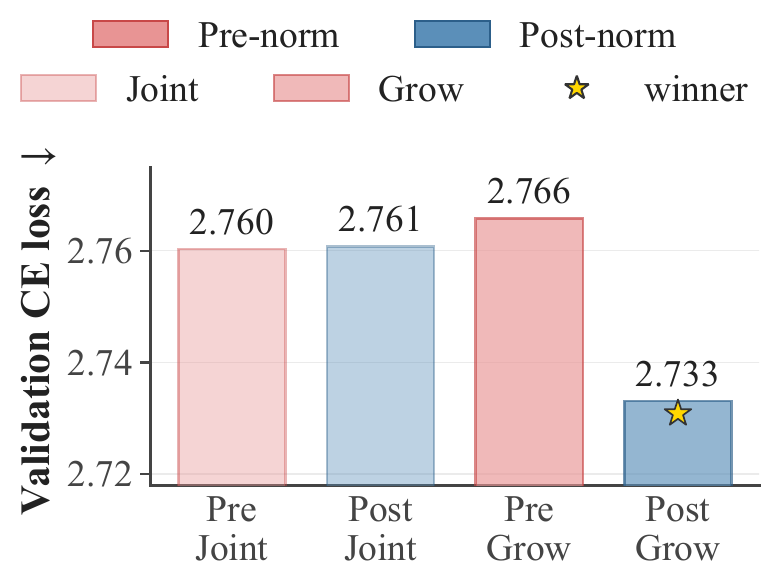}
\caption{Placement--curriculum interaction.}
\label{fig:main}
\end{subfigure}\hfill
\begin{subfigure}[t]{0.48\textwidth}
\centering
\includegraphics[width=\linewidth]{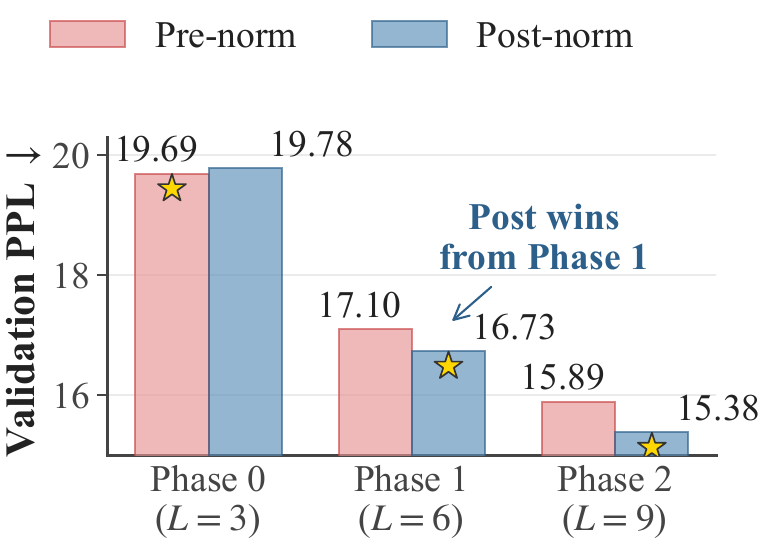}
\caption{Phase-wise crossover under curriculum growth.}
\label{fig:stages}
\end{subfigure}
\caption{Main placement--curriculum results. The pre--post ranking changes under curriculum growth, and the crossover appears after the initial shallow phase.}
\label{fig:main-results}
\end{figure*}

\subsection{Experimental Setup}
\label{sec:setup}

\paragraph{Teacher, Student, and Data.}
We use the base Qwen3-8B model~\cite{DBLP:journals/corr/abs-2505-09388} as the teacher and train a block-stack student with $K{=}3$ blocks and $L{=}3$ decoder layers per block, giving $S{=}9$ trainable layers in total.
Following the block-stack distillation setting~\cite{DBLP:journals/corr/HintonVD15,DBLP:conf/emnlp/JiaoYSJCL0L20}, the student copies the teacher embedding $\mathbf{E}$ and language-modeling head $\mathbf{W}_{\mathrm{lm}}$. The embedding remains frozen, while the language-modeling head is optimized jointly with decoder layers initialized using the default Qwen3 module-wise scheme with initializer range $0.02$.
All students are distilled on the FineWeb-Edu 10BT split~\cite{DBLP:conf/nips/PenedoKALMRW024}.
All reported primary runs use seed 42, a global batch size of $64$, and a maximum sequence length of $8192$ across all configurations and phases.

\paragraph{Compared Configurations.}
We compare four configurations formed by crossing normalization placement and depth-introduction protocol: \{pre-norm, post-norm\}$\times$\{joint, grow\}.
In joint training, all blocks are active from the first update and are trained end-to-end.
In curriculum growth, blocks are appended sequentially following Algorithm~\ref{alg:grow}: each phase activates one additional block, carries over the previously trained weights, and introduces a newly initialized block when depth increases.
Within each paired placement comparison, the teacher, student architecture, data order, objective, optimizer configuration, and initialization protocol are matched. Joint and grow then follow their respective activation, duration, replay, and reset schedules.

\paragraph{Training Budget.}
Each grow phase runs for $50{,}000$ steps.
Because sequences are variable-length and most documents fall well below the $8192$-token cap, each phase processes approximately $3.11$B non-padding tokens rather than the fully-packed upper bound.
The three grow phases consume the same data shard, so grow sees $3.11$B unique tokens repeated across phases ($9.34$B total).
Joint training runs $50{,}000$ steps over the same $3.11$B-token shard.
At every phase transition, the optimizer state is reset and the learning-rate schedule is restarted with warmup.
Because growth activates fewer layers in early phases, we report a unique-token-matched primary comparison and active-layer-token controls; Appendix Table~A2 summarizes the budget for each configuration.
The primary setting matches unique-token exposure, while the compute proxy matches student active-layer tokens, computed as training tokens multiplied by the number of active decoder layers.

\paragraph{Evaluation Metrics.}
The primary metric is validation cross-entropy (CE) loss $\downarrow$ on a fixed held-out set of $4{,}833$ documents (approximately $4.98$M tokens), evaluated every $2{,}000$ steps, with perplexity (PPL) $\downarrow$ reported as an equivalent scale.
For mechanism analysis, we measure block-boundary residual RMS across grow checkpoints and inspect validation-loss trajectories around phase transitions.

\subsection{Main Results}
\label{sec:main}

For \textbf{RQ1}, we evaluate whether the relative behavior of pre-norm and post-norm changes across depth-introduction protocols.
We compare the two placements under matched architecture, data, objective, and unique-token exposure; grow performs more optimization steps, examined through active-layer-token controls.

\paragraph{Placement--Curriculum Interaction.}
Table~\ref{tab:main} and Figure~\ref{fig:main} report the primary comparison.
Under joint training, pre-norm and post-norm reach nearly identical validation losses, with CE values of $2.7603$ and $2.7607$, respectively.
Under curriculum growth, the ranking changes: post-norm reduces validation CE from $2.7658$ to $2.7330$ compared with pre-norm.
This yields $\Delta_{\text{joint}}=+0.0004$ and $\Delta_{\text{grow}}=-0.0328$, where $\Delta=\text{Post}-\text{Pre}$. The joint gap indicates how far the two placements drift apart when depth is held fixed, and the grow gap is roughly eighty times larger.
The resulting interaction, $\Delta_{\text{grow}}-\Delta_{\text{joint}}=-0.0332$, quantifies the observed shift toward post-norm under curriculum growth in this distillation setting.

\begin{table}[t]
\centering
\small
\begin{tabular}{lrr}
\toprule
 & \multicolumn{2}{c}{Unique-token-matched} \\
\cmidrule(lr){2-3}
 & Joint & Grow \\
\midrule
\preRow{Pre-norm}  & \best{2.7603} & \base{2.7658} \\
\postRow{Post-norm} & \base{2.7607} & \best{2.7330} \\
\addlinespace
$\Delta$ (Post$-$Pre) & $+$0.0004 & $-$0.0328 \\
Interaction           & \multicolumn{2}{c}{$-$0.0332} \\
\bottomrule
\end{tabular}
\caption{Primary placement--curriculum interaction.
Validation CE loss $\downarrow$ under unique-token-matched budgets.
$\Delta=\text{Post}-\text{Pre}$, so negative values indicate a post-norm advantage.}
\label{tab:main}
\end{table}

\paragraph{Phase-wise Crossover.}
The per-phase results further show where the ranking changes during curriculum growth.
As shown in Figure~\ref{fig:stages}, pre-norm is slightly better in the shallow Phase~0, where the student contains only one block.
After additional blocks are appended, post-norm becomes better in both Phase~1 and Phase~2.
This crossover matches the expected shift from optimizing a shallow block to conditioning the boundary distribution passed to appended blocks.
We next use single-block and active-layer-token controls to test whether shallow-block quality or longer optimization can explain the pattern.

\subsection{Controlled Analysis}
\label{sec:controls}

For \textbf{RQ2}, we examine whether the post-norm advantage under curriculum growth can be explained by simpler alternatives.
We test two possibilities under the same held-out evaluation: post-norm may form stronger individual blocks, or curriculum growth may benefit from additional student-decoder active compute.

\paragraph{Single-Block Control.}
We first test whether post-norm is already stronger when training a single block in isolation.
Table~\ref{tab:single} reports Phase~0 results, where the student contains only one block ($L{=}3$).
Pre-norm is marginally ahead on CE, KD loss, and PPL, with all gaps below $0.01$ in CE/KD, which rules out post-norm forming a better shallow block as the source of its grow advantage; instead, the advantage appears only after additional blocks are appended, matching the phase-wise crossover in Figure~\ref{fig:stages}.

\begin{table}[t]
\centering
\small
\begin{tabular}{lrrr}
\toprule
Norm & CE $\downarrow$ & KD $\downarrow$ & PPL $\downarrow$ \\
\midrule
\preRow{Pre}  & \best{2.9802} & \best{3.0155} & \best{19.6900} \\
\postRow{Post} & \base{2.9846}  & \base{3.0183}  & \base{19.7800}  \\
\bottomrule
\end{tabular}
\caption{Single-block Phase~0 results.
CE, KD loss, and PPL are reported for one-block pre-norm and post-norm students.}
\label{tab:single}
\end{table}

\paragraph{Active-Layer-Token Control.}
We next test whether additional student-decoder active compute explains the grow advantage.
Because curriculum growth traverses three phases while joint training uses one full-depth phase, we extend post-joint to matched active-layer budgets; these controls use phase-aligned 50k-step restarts at fixed depth and continue onto fresh data rather than replaying the grow shard.
Table~\ref{tab:longer} shows that post-joint improves from $2.7607$ to $2.7518$ when matched to grow on student active-layer tokens, and to $2.7428$ when matched to grow on total training tokens while using more student-decoder active compute. Both nonetheless remain worse than post-grow ($2.7330$), despite the additional compute.
This suggests that the post-norm grow result is not reducible to additional student active-layer compute under the evaluated schedules.

\begin{table}[t]
\centering
\small
\begin{tabular}{lcr}
\toprule
Run & Matches grow on & CE $\downarrow$ \\
\midrule
\postRow{Post-Joint}        & Unique tokens & \base{2.7607} \\
\postRow{Post-Joint-long}   & Active-layer tokens & \base{2.7518} \\
\postRow{Post-Joint-longer} & Total tokens & \base{2.7428} \\
\addlinespace
\bottomrule
\end{tabular}
\caption{Post-joint budget controls.
Validation CE loss $\downarrow$ under three matching criteria.}
\label{tab:longer}
\end{table}

\begin{figure}[t]
\centering
\includegraphics[width=\columnwidth]{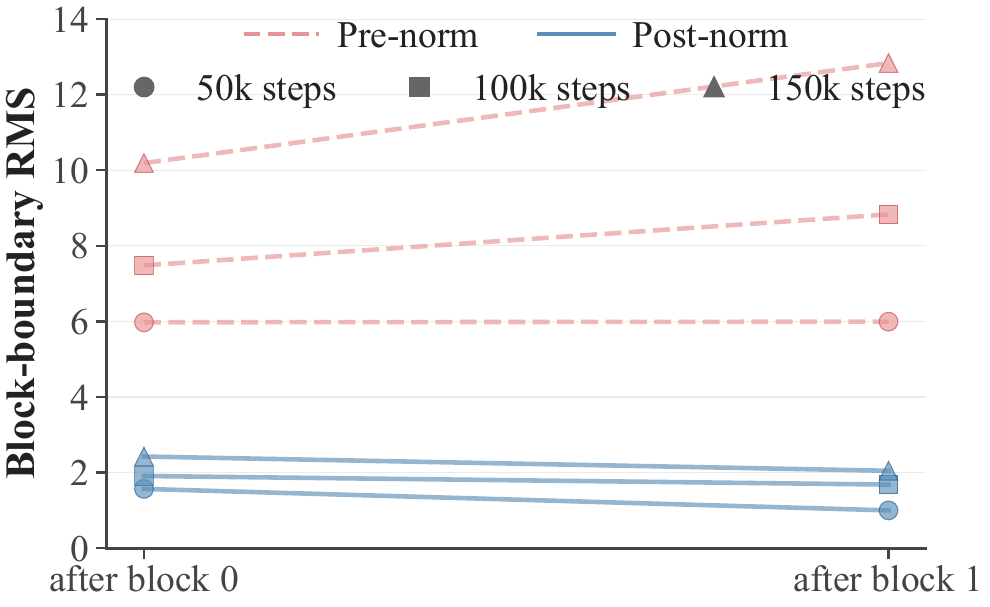}
\caption{Block-boundary RMS across growth phases.}
\label{fig:rms-main}
\end{figure}

\subsection{Boundary-Scale Diagnostics}
\label{sec:mech-exp}

For \textbf{RQ3}, we test whether block-boundary dynamics support forward conditioning as a contributing mechanism.
At each depth transition, a newly appended block receives the residual stream produced by the prefix trained in the preceding phase.
We examine residual-stream scale and token composition, appended-block contributions, transition-localized loss, and the freeze control, which together characterize how the boundary distribution evolves as depth grows.

\begin{figure*}[t]
\centering
\begin{subfigure}[t]{0.48\textwidth}
\centering
\includegraphics[width=\linewidth]{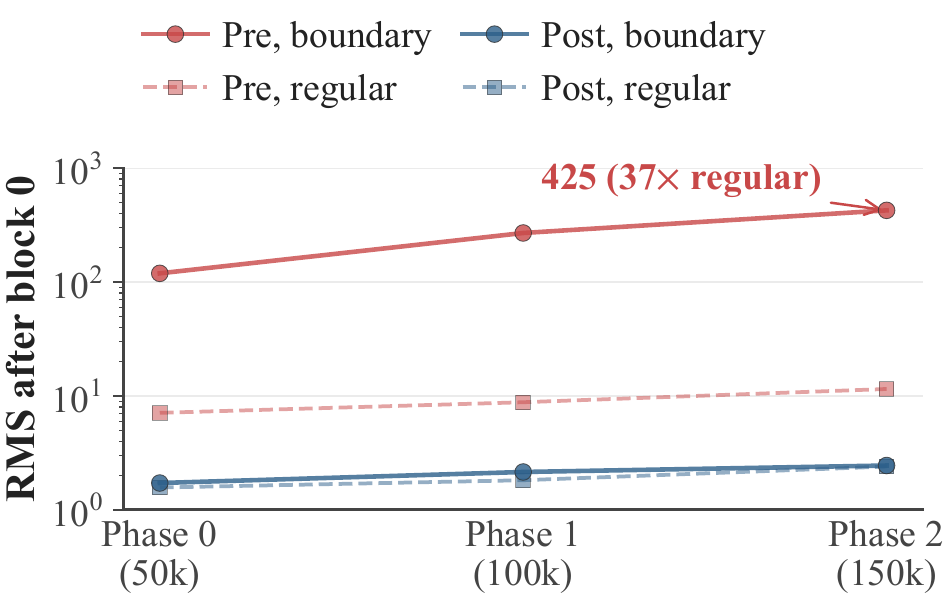}
\caption{Token-type attribution of scale drift.}
\label{fig:rms-attribution}
\end{subfigure}\hfill
\begin{subfigure}[t]{0.48\textwidth}
\centering
\includegraphics[width=\linewidth]{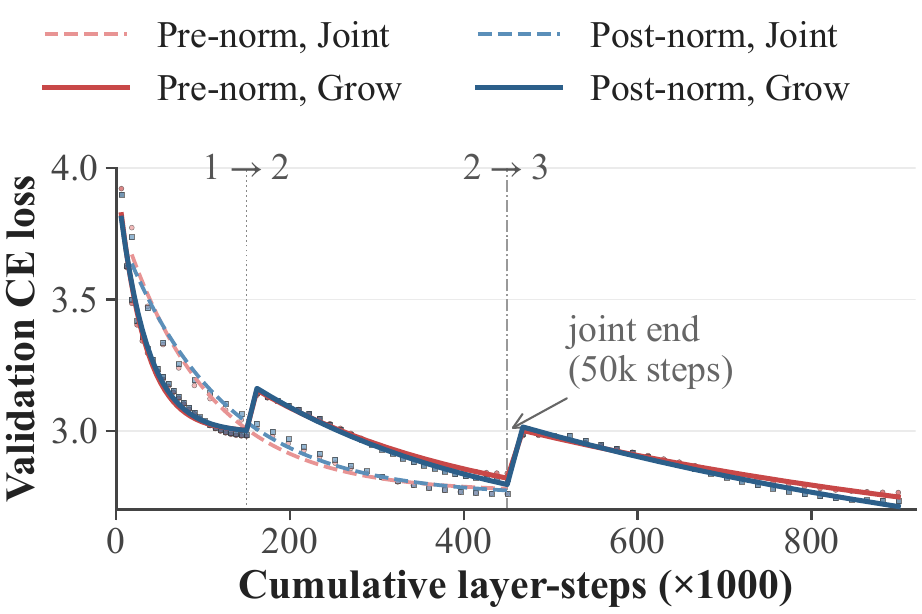}
\caption{Validation-loss trajectories and grow transitions.}
\label{fig:curves}
\end{subfigure}
\caption{Forward-conditioning diagnostics. Post-norm exhibits limited structural-token scale drift, and the placement gap emerges after block-appending transitions.}
\label{fig:mechanism-main}
\end{figure*}

\paragraph{Block-Boundary Scale Drift.}
Figure~\ref{fig:rms-main} shows that post-norm keeps block-boundary RMS controlled, whereas pre-norm drifts as depth grows: after block~0, the per-token RMS standard deviation rises from $5.81$ to $19.98$ under pre-norm but remains approximately $0.07$--$0.19$ under post-norm. On the final-phase diagnostic batch (Appendix Figure~A1), pre-norm shows a heavy-tailed distribution (kurtosis $>2400$) versus a far less heavy-tailed post-norm distribution.

\paragraph{Token-Type Attribution of Scale Drift.}
The extreme kurtosis suggests that pre-norm's drift is driven by a small subset of tokens rather than a uniform shift.
Figure~\ref{fig:rms-attribution} localizes pre-norm's drift to structural tokens: document-boundary RMS grows from $119$ to $425$ ($37\times$ regular-token RMS at 150k), while regular-token RMS grows only from $7.1$ to $11.5$; post-norm keeps both below $2.5$ ($2.45$ vs.\ $2.40$), so the drift is concentrated at structural-token massive activations rather than a uniform global shift.

\paragraph{Block-Removal Sensitivity and Representation Change.}
On a fixed $2\times 2048$-token diagnostic batch (seed 42), we bypass each entire block and measure $\Delta\mathrm{CE}_l = \mathrm{CE}(\text{skip block } l) - \mathrm{CE}(\text{full model})$, together with angular distance $\arccos(\mathrm{cosine}(\mathbf{h}_{\mathrm{in}},\mathbf{h}_{\mathrm{out}}))/\pi$.
For a post-norm three-layer block, bypass removes its attention and MLP modules together with six post-residual RMSNorm operations; the model entry norm remains outside the skipped block. Thus, $\Delta\mathrm{CE}$ measures whole-block removal sensitivity rather than the attention and MLP contribution alone.

Under pre-norm growth, bypassing the final block changes CE by only $0.02$, and its angular distance is $0.031$, indicating a near-identity mapping.
Removing any post-grow block instead raises CE by at least $2.38$, with angular distances above $0.34$; neither joint model has a near-identity final block. Post-joint places low removal sensitivity on block~0 ($\Delta\mathrm{CE}=0.55$) but high sensitivity on block~2 ($4.00$), showing that uneven allocation alone does not imply the pre-grow failure pattern.
Thus, scale drift and a nearly identity-mapped final block coincide under pre-norm growth, while post-norm retains controlled scales and nontrivial representation changes across blocks.

\begin{table}[t]
\centering
\small
\setlength{\tabcolsep}{3pt}
\begin{tabular}{@{}llrrr@{}}
\toprule
Run & Metric & Block 0 & Block 1 & Block 2 \\
\midrule
\preRow{Pre-Grow}  & $\Delta$CE$\uparrow$ & 8.13 & 2.36 & 0.02 \\
                    & Angular dist. & 0.499 & 0.310 & 0.031 \\
\addlinespace
\postRow{Post-Grow} & $\Delta$CE$\uparrow$ & 8.32 & 3.74 & 2.38 \\
                    & Angular dist. & 0.483 & 0.365 & 0.342 \\
\addlinespace
\preRow{Pre-Joint}  & $\Delta$CE$\uparrow$ & 8.05 & 1.47 & 2.34 \\
                    & Angular dist. & 0.485 & 0.245 & 0.360 \\
\addlinespace
\postRow{Post-Joint} & $\Delta$CE$\uparrow$ & 0.55 & 0.99 & 4.00 \\
                     & Angular dist. & 0.395 & 0.166 & 0.437 \\
\bottomrule
\end{tabular}
\caption{Block-removal sensitivity and representation change.
$\Delta$CE includes removal of internal normalization; angular distance measures input--output change.
Only Pre-Grow has a near-identity final block ($\Delta$CE$\approx 0$, angular $\approx 0.03$).}
\label{tab:block-contrib}
\end{table}

\paragraph{Transition-Localized Loss Divergence.}
Figure~\ref{fig:curves} shows that the placement gap emerges after blocks are appended. Transition-aligned loss reveals that post-norm incurs larger immediate increases at both transitions, with much slower recovery after the first and only a small delay after the second (Appendix Table~A4), whereas pre-norm changes more smoothly.
The smooth pre-norm transition is consistent with a smaller initial perturbation, yet its final block is nearly identity-mapped; post-norm incurs a larger adjustment but develops nontrivial block changes. Thus, boundary-scale control should not be interpreted as a smoother immediate transition: transition smoothness and final block utilization need not coincide in general.

\paragraph{Freeze Control.}
Finally, we freeze completed blocks and update only the newly appended block under matched budgets and transition policies. Freezing worsens both placements by approximately $0.144$ CE but leaves the gap nearly unchanged ($-0.0328$ vs. $-0.0332$), so retraining earlier blocks does not account for the placement gap (Appendix Table~A3). These diagnostics observe rather than intervene on the boundary distribution, and together they provide converging evidence for boundary-scale conditioning.

\section{Conclusion}

Pre-norm and post-norm are nearly tied under joint training, whereas post-norm is better under curriculum depth growing. Single-block, compute-matched, freeze, and boundary diagnostics associate this crossover with controlled boundary scales, ruling out shallow-block quality, active-layer compute, or retraining alone. The results therefore motivate treating normalization placement and depth curriculum as coupled rather than independent design choices.

\bibliography{aaai2027}

\clearpage
\appendix
\setcounter{table}{0}
\setcounter{figure}{0}
\renewcommand{\thetable}{A\arabic{table}}
\renewcommand{\thefigure}{A\arabic{figure}}
\section{Appendix A: Training and Compute Protocol}

\subsection{A.1 Detailed Hyperparameter Settings}
\label{app:hyper}

Table~\ref{tab:hyperparams} lists shared hyperparameters. Training uses AdamW with cosine decay and warmup.

\subsection{A.2 Budget and Compute Protocol}
\label{app:budget}

Because curriculum growth activates fewer layers in early phases, step-matched budgets are not compute-fair. We use a unique-token-matched primary comparison and additional post-joint active-layer-token controls:

\begin{itemize}
    \item \textbf{Unique-token-matched}: joint and grow see the same \emph{unique} tokens. This compares quality at equal data coverage, although grow uses more active-layer compute across three phases.
    \item \textbf{Active-layer-token control}: post-joint is extended to the active-layer-token budget of grow, i.e.\ trained tokens multiplied by the number of active decoder layers at each step. We additionally report a $3\times$ post-joint run beyond that proxy-matched budget.
\end{itemize}

Compute can be approximated by active-layer tokens:
\[
\text{active-layer tokens} = \sum_{\text{step }t} (\text{tokens at }t) \cdot (\text{active layers at }t).
\]
\begin{table}[ht]
\centering
\footnotesize
\begin{tabular}{ll}
\toprule
Hyperparameter & Value \\
\midrule
Optimizer & AdamW \\
Optimizer $\beta_1, \beta_2$ & $0.9,\ 0.95$ \\
Optimizer $\epsilon$ & $1\text{e-}8$ \\
Weight Decay & $0$ \\
Peak Learning Rate & $1\text{e-}4$ \\
Minimum Learning Rate & $1\text{e-}6$ \\
Learning Rate Schedule & Cosine decay with warmup \\
Warmup Steps & $0.015\times 50{,}000 = 750$ \\
Gradient Clipping & $0.5$ (global L2 norm) \\
Training Precision & BF16 \\
Global Batch Size & $64$ \\
Sequence Length & $8192$ (max) \\
Evaluation Interval & $2{,}000$ steps \\
Distillation Weight $\alpha$ & $0.7$ \\
Language Modeling Weight $\gamma$ & $0.3$ \\
Temperature $\tau$ & $1.0$ \\
Dropout & $0$ \\
Attention Dropout & $0$ \\
Seed & $42$ \\
\bottomrule
\end{tabular}
\caption{Hyperparameters shared across the primary $2\times2$ configurations. Controls additionally vary training duration or trainable blocks as stated in their protocols.}
\label{tab:hyperparams}
\end{table}

\begin{table}[ht]
\centering
\small
\setlength{\tabcolsep}{2pt}
\begin{tabular}{@{}lccrr@{}}
\toprule
Config & Phases & Layers & Total tok. & Unique tok. \\
\midrule
Joint-50k & 1 & 9 & 3.11B & 3.11B \\
Grow-3$\times$50k & 3 & 3$\to$6$\to$9 & 9.34B & 3.11B \\
Joint (proxy-matched) & 1 & 9 & 6.22B & 6.22B \\
Joint-150k & 1 & 9 & 9.34B & 9.34B \\
\bottomrule
\end{tabular}
\caption{Training budgets. Grow reuses one fixed shard, whereas longer joint controls continue onto fresh data; proxy-matched joint matches student active-layer tokens, and Joint-150k is the $3\times$ stress test.}
\label{tab:budget}
\end{table}

\subsection{A.3 Phase Transition Optimization Reset}
\label{app:reset}

At each curriculum growth transition, we reset the optimizer state and restart the learning-rate schedule with warmup for two reasons:

\begin{enumerate}
    \item \textbf{Depth increase.} Moving from $p$ to $p+1$ blocks introduces newly initialized weights. Restarting the LR schedule with warmup updates these layers smoothly instead of exposing them to stale momentum.
    \item \textbf{Stale gradient statistics.} Resetting Adam momentum prevents statistics accumulated in the shallower phase from biasing the deeper network.
\end{enumerate}

Thus, each phase inherits model weights but re-estimates optimizer statistics rather than carrying shallower-phase momentum across successive transitions.

\section{Appendix B: Additional Analyses}

\subsection{B.1 Freeze Ablation: Probing Forward Conditioning}
\label{app:freeze}

\paragraph{Setup.} The main grow protocol retrains all active blocks in every phase. A natural alternative is to \emph{freeze} each block once its phase completes, so that phase $p$ trains only the newly appended block $p$ while blocks $0,\dots,p-1$ are held fixed. Freeze and retrain differ in two ways: (i) freeze gives weaker absolute performance because earlier blocks cannot adapt to the deeper stack, and (ii) under freeze the input distribution to a newly appended block is \emph{strictly fixed} (it is the exact output of frozen blocks), which sharpens the forward-conditioning channel. If retraining primarily compensated for pre-norm's boundary drift, the post--pre gap would be expected to amplify under freeze. Instead, the observed gap is essentially unchanged. However, freeze does not fully isolate the forward channel: the newly appended block is still trained, so its backward-gradient dynamics at the append step still differ across placements. The result is therefore compatible with a forward-conditioning contribution, but does not exclude a backward-at-append contribution.

\paragraph{Protocol.} The freeze variant follows the grow protocol of the main paper exactly, except that at the start of phase $p\!+\!1$ the parameters of blocks $0,\dots,p$ are frozen (no gradient, no optimizer state); only block $p\!+\!1$ receives updates. Step budgets, LR schedule, data shards, and the reset-at-transition policy are identical to grow. Pre-norm and post-norm use the same paired seed and initialization as the main runs.

\paragraph{Result.} Table~\ref{tab:freeze} reports validation CE at 150k. Freeze is uniformly worse than retrain by $\approx 0.144$ CE for both placements, showing that retraining earlier blocks is beneficial in absolute terms. The placement gap, however, is essentially unchanged: $\Delta_{\text{grow}}=-0.0328$ versus $\Delta_{\text{freeze}}=-0.0332$, a difference of $-0.0004$ ($\approx 1.2\%$ of the gap). On PPL the gap is wider under freeze ($-0.60$ vs $-0.51$), but PPL is the exponential of CE, so this amplification is consistent with an unchanged CE gap.

\begin{table}[ht]
\centering
\small
\begin{tabular}{lrr}
\toprule
 & Grow (retrain) & Freeze \\
\midrule
\preRow{Pre-norm}  & 2.7658 & 2.9104 \\
\postRow{Post-norm} & 2.7330 & 2.8772 \\
\addlinespace
$\Delta$ (Post$-$Pre) & $-$0.0328 & $-$0.0332 \\
$\Delta_{\text{freeze}}-\Delta_{\text{grow}}$ & \multicolumn{2}{c}{$-$0.0004} \\
\bottomrule
\end{tabular}
\caption{Freeze ablation. Validation CE loss $\downarrow$ at 150k under the unique-token-matched budget (grow and freeze both traverse three 50k-step phases over the same shard). Grow reproduces the main interaction (retrain); freeze holds previously appended blocks fixed. $\Delta=\text{Post}-\text{Pre}$ (negative $=$ post wins). The placement gap is essentially unchanged across retrain/freeze ($|\Delta|$ differs by $0.0004$).}
\label{tab:freeze}
\end{table}

\paragraph{Interpretation.} The result is consistent with the forward-conditioning account, with two caveats. First, the absolute freeze penalty ($\approx 0.144$ CE) is nearly identical across placements, so a retraining-only explanation for post-norm's relative advantage is unlikely under this control. Second, the gap's near-invariance to retrain/freeze is consistent with the RQ3 RMS evidence: pre-norm's boundary drift is already present in the outputs of the prefix at the append step, whether or not those blocks are retrained later. The result therefore weighs against the stronger claim that retraining actively \emph{masks} pre-norm's drift, and suggests that later joint updates do not recover the full append-time cost. We emphasize what freeze does \emph{not} establish: because the newly appended block remains trainable, its backward-gradient dynamics at the append step still differ by placement, so the gap's stability is also compatible with a backward-at-append channel. Forward conditioning is therefore a candidate explanation for post-norm's grow advantage, but isolating it as the sole cause would require an intervention that changes the forward boundary distribution without changing backward updates.

\subsection{B.2 Mechanism Diagnostics}
\label{app:mechanism}

\begin{figure}[t]
\centering
\includegraphics[width=\columnwidth]{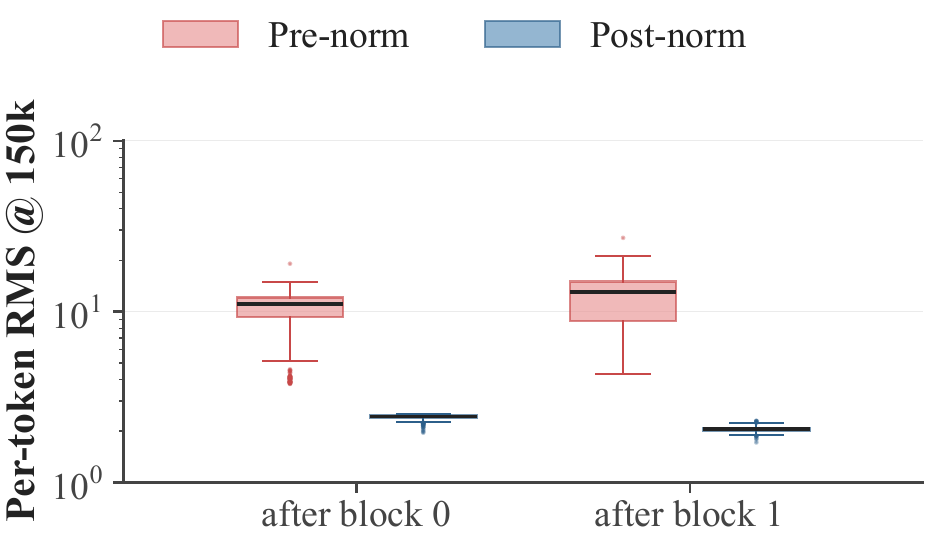}
\caption{Final-phase per-token RMS distributions.}
\label{fig:mechanism-appendix-b}
\end{figure}

\begin{figure*}[t]
\centering
\includegraphics[width=\textwidth]{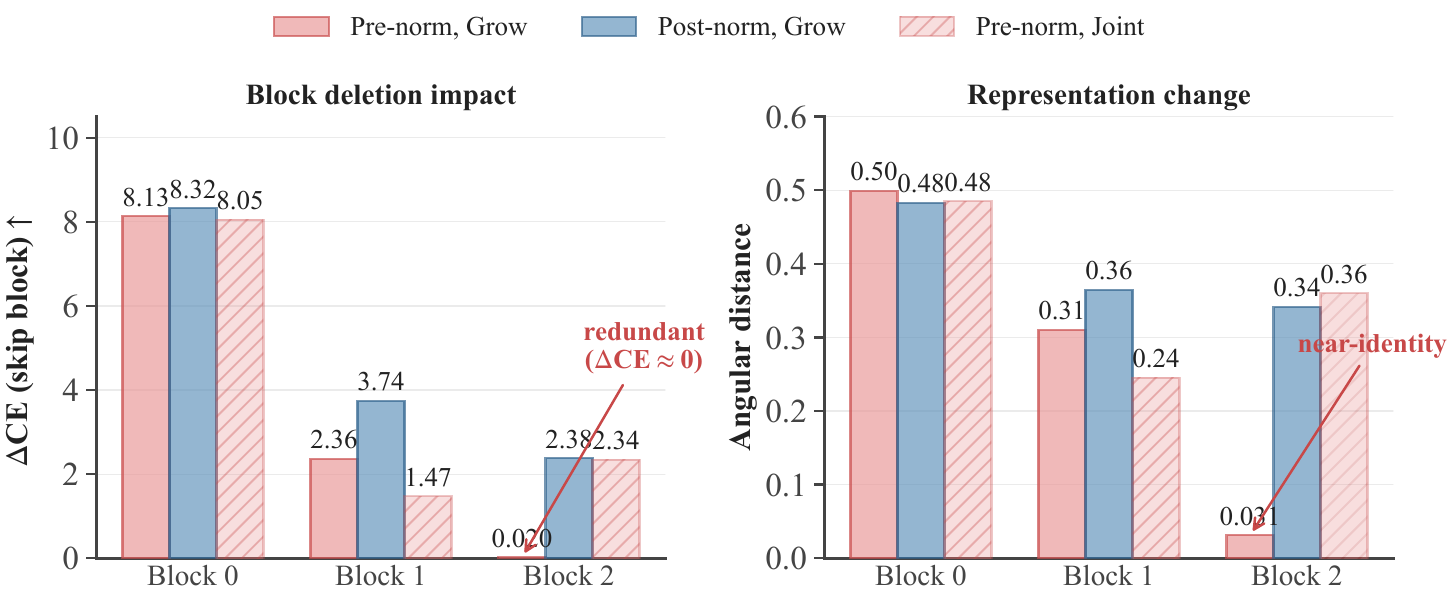}
\caption{Whole-block removal sensitivity and representation change. Whole-block bypass also removes internal normalization operations.}
\label{fig:mechanism-appendix-c}
\end{figure*}

The main paper reports the primary interaction and phase-wise crossover in Figure~3, and the block-boundary RMS trend, token-type attribution, and transition-localized loss trajectories in Figures~4 and~5. Figures~\ref{fig:mechanism-appendix-b} and~\ref{fig:mechanism-appendix-c} complement those aggregate trends with the final-checkpoint RMS distribution and per-block deletion behavior.

\begin{table}[t]
\centering
\small
\setlength{\tabcolsep}{3pt}
\begin{tabular}{@{}lrr@{}}
\toprule
Metric & 50k (Pre/Post) & 100k (Pre/Post) \\
\midrule
Loss jump & 0.0226 / 0.2500 & 0.0230 / 0.2847 \\
AUC$_{1\mathrm{k}}$ & 0.0694 / 0.1820 & 0.0688 / 0.2059 \\
AUC$_{5\mathrm{k}}$ & 0.1217 / 0.1464 & 0.1277 / 0.1774 \\
Recovery steps & 101 / 19,914 & 19,054 / 19,426 \\
\bottomrule
\end{tabular}
\caption{Transition-aligned training-loss response at the 50k and 100k block-appending steps. Jump and AUC are measured relative to the pre-transition baseline; recovery is the first step at which a 200-step smoothed loss returns to that baseline. Post-norm pays a larger immediate adjustment cost despite its better final grow loss.}
\label{tab:transition-aligned}
\end{table}

\subsection{B.3 Additional Protocol Checks}

\paragraph{Data-Iterator Reset.} Within curriculum growth, we compared resetting the data iterator at phase boundaries with continuing from its current position. Both variants draw from the same fixed 3.11B-token shard and wrap within that shard after reaching its end; a phase transition never switches to new data. They produce similar final losses and the same placement ranking. The main results use the no-reset variant, so a phase change continues from the current offset in this repeating cycle rather than explicitly rewinding it, while unique-token coverage remains 3.11B. The longer joint controls instead continue onto fresh data, so their total- and unique-token counts coincide.

\paragraph{Joint Learning-Rate Restarts.} The 100k and 150k post-joint controls in Table~\ref{tab:budget} keep all nine decoder layers active and restart the 50k-step learning-rate schedule at the same boundaries used by grow. Each restart produces a transient loss increase followed by renewed descent, but the subsequent decrease is slower than the corresponding curriculum-growth trajectory. Their endpoint CE values improve to $2.7518$ and $2.7428$, respectively, yet remain above post-grow at $2.7330$; LR restarting alone therefore does not reproduce the grow trajectory.

\section{Appendix C: Implementation Details}

\paragraph{Model and Systems.}
The Qwen3-8B-Base teacher has 36 decoder layers. The nine-layer student uses hidden size 4096, 32 attention heads, 8 key--value heads, and intermediate size 12288. It contains approximately 3.09B parameters including the teacher-initialized embedding and language-modeling head, 2.47B when excluding the head but retaining the embedding, and 1.85B in the decoder stack alone. The embedding is frozen, whereas the language-modeling head is trained. Training uses FSDP over 8 GPUs, while teacher inference uses a separate 8-GPU pool; gradient checkpointing and MLP recomputation are enabled.

\paragraph{Data Processing and Evaluation.}
Examples are tokenized independently, truncated to 8192 tokens, and right-padded to the longest sequence in each batch; multiple documents are not packed into one sequence. Padding is excluded from the token accounting and loss. The fixed validation set contains 4,833 documents and approximately 4.98M tokens, and evaluation is run every 2,000 optimization steps. Main tables report the scheduled endpoints rather than selecting the best validation checkpoint.

\end{document}